\documentclass[runningheads]{llncs}

\usepackage[mobile]{eccv}

\usepackage{eccvabbrv}

\usepackage{graphicx}
\usepackage{booktabs}

\usepackage[accsupp]{axessibility}  

\usepackage{hyperref}

\usepackage{orcidlink}

\usepackage{xcolor} 
\usepackage{graphicx}
\usepackage{amsmath}
\usepackage{amssymb}
\usepackage{booktabs}
\usepackage{graphicx}
\usepackage{subcaption}
\usepackage{multirow}
\usepackage{spverbatim}
\usepackage{afterpage}
\usepackage{wrapfig}
\newcommand{\model}{NoisyCLIP\xspace}
\newcommand{\modelsiglip}{NoisySigLIP\xspace}

\begin{document}

\title{Early Estimation of Language to Latent Alignment in Diffusion Models}

\titlerunning{Estimation of Alignment in Diffusion Models}

\author{Vasco Ramos\inst{1}\orcidlink{0000-0002-2453-3288} \and
Regev Cohen\inst{2}\orcidlink{0000-0002-2833-8965} \and
Idan Szpektor\inst{2}\orcidlink{0009-0002-2746-4027}
\and
Joao Magalhaes\inst{1}\orcidlink{0000-0001-6290-5719}}

\authorrunning{V.~Ramos et al.}

\institute{NOVALINCS, NOVA University of Lisbon, Portugal \and
Google Research \\
\email{vcc.ramos@campus.fct.unl.pt}\\
\url{https://novasearch.github.io/noisyclip}}

\maketitle

\begin{abstract}
Conditional diffusion models frequently suffer from language-image misalignments. Due to the ambiguity of intermediate noise corrupted latents, assessing prompt adherence currently requires completing the entire sampling trajectory. This late-stage evaluation incurs even higher computational costs during test-time scaling strategies, such as Best-of-N (BoN) sampling, as all misaligned trajectories must finish generation before being discarded. To tackle this, we propose \model, a noise-aware twin-tower model that enables early language-to-latent alignment estimation. By learning a vision encoder on noise-corrupted latents, we allow the model to "see" through the ambiguity of intermediate diffusion steps. To facilitate this training, we investigate noise-data augmentation sampling strategies and introduce two new benchmark datasets: Noisy-Conceptual-Captions and Noisy-GenAI-Bench. When applied as an early-stopping criterion for BoN, \model at half cost matches or beats frozen CLIP at full cost. Ultimately, this transforms alignment assessment from an expensive final check into a continuous monitoring tool, drastically reducing compute costs without sacrificing semantic fidelity.
\end{abstract}

\section{Introduction}
The advances in image generation achieved by diffusion models~\cite{ddpm,stablediffusion,sdxl,imagen,imagen3} have opened exciting possibilities for generating realistic images from textual prompts. Reverse diffusion, coupled with cross-attention mechanisms, guides iterative removal of noise, shaping a final image that adheres to the input textual prompt. However, reverse diffusion image generation is a costly process,  and there is no inherent guarantee that it will follow the optimal denoising trajectory~\cite{a-star}, often failing to produce an image that fully adheres to the prompt.

\begin{figure}[t]
    \centering
    \includegraphics[width=0.94\linewidth]{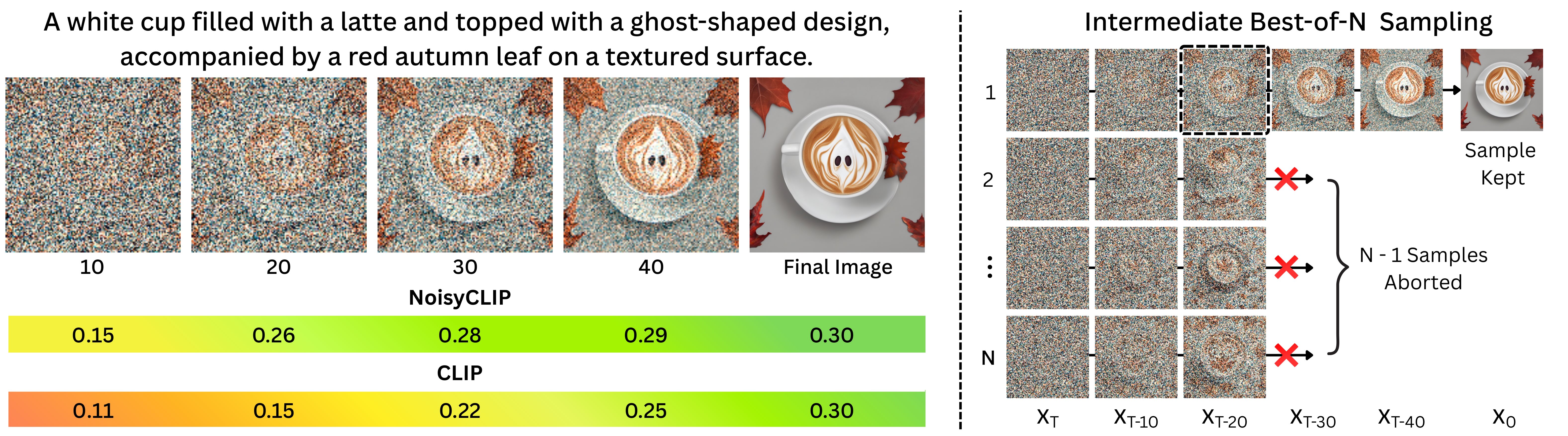}
    \caption{\textbf{Left:} While standard CLIP scores require near-complete generation to stabilize, \model accurately predicts final image quality at early denoising steps. \textbf{Right:} We leverage this early evaluation for intermediate Best-of-N sampling. Evaluating multiple candidates early (e.g., $X_{T-20}$) terminates unpromising trajectories, significantly reducing compute overhead while allowing the highest-scoring candidate to reach $X_0$.}
    \label{fig:initial_figure}
    \vspace{-5mm}
\end{figure}

Current methodologies for measuring alignment between textual prompts and images mainly focus on assessing alignment of the output image~\cite{clipscore,vqascore,pick-a-pic}.
Nonetheless, the iterative nature of diffusion models raises a fundamental challenge: \textbf{accurately measuring language-to-latent alignment across different intermediate diffusion timesteps}.
The core difficulty in providing real-time feedback lies in the nature of the latent space itself during the early stages of generation, where representations are dominated by Gaussian noise, making traditional vision-language encoders ineffective~\cite{cosed}. 
To address this, we propose NoisyCLIP, a noise-aware twin-tower model designed to actively measure language-to-latent alignment during the denoising process, Figure~\ref{fig:initial_figure}.
By learning a vision encoder over noise-corrupted representations, we equip the model to accurately interpret intermediate diffusion stages, transforming alignment assessment from a final check into a continuous monitoring tool.

Learning such language-to-latent estimator is a non-trivial task.
Although noise inversion techniques exist~\cite{prompt_tuning_inversion,cfg_google}, there are no obvious data resources or sampling strategies that can reliably provide the intermediate noise training sequences.
This leads to our second core challenge: \textbf{identifying the training objectives and data augmentation strategies necessary to stabilize a latent-alignment estimator}.
To address this critical gap, we investigate noise-data augmentation sampling strategies for training and assessing alignment of intermediate latent representations in reverse diffusion processes.

Finally, to validate the system-level utility of our approach, we address a final challenge: \textbf{establishing an alignment-based early-stopping criterion that significantly reduces Best-of-N computational costs while maintaining full-term image-text fidelity.}
Evaluating solely on the final image results in wasted compute time since misaligned generations must complete the full process before being discarded. This inefficiency is amplified in test-time scaling paradigms such as Best-of-N (BoN) selection~\cite{inference_time_scaling,genai-bench,xie2025sana}, where having multiple images incurs a substantial collective computational cost. 

Our main contributions are threefold: we propose \model, a novel noise-aware architecture capable of measuring language-to-latent alignment during the highly ambiguous intermediate stages of diffusion; we introduce two new benchmark datasets, Noisy-Conceptual-Captions and Noisy-GenAI-Bench, containing full sequences of intermediate latents to evaluate these alignments; and we pioneer an early-stopping Best-of-N strategy that cuts computational costs by 50\% while retaining 98\% of post-hoc CLIP performance across multiple architectures.

\section{Related Work}

\paragraph{Text-to-image Generation.} These models are trained to learn associations between linguistic and visual concepts~\cite{diffusion_survey}. 
Despite previous efforts~\cite{gans,vae}, reverse diffusion models, including DALL-E 2~\cite{dalle2}, Imagen~\cite{imagen}, and Stable Diffusion~\cite{sd}, have emerged as the predominant approach~\cite{sd,sdxl,imagen3,sd3,dit}. Although early diffusion models~\cite{ddpm} lacked control over the generation, advances in controllability and quality have been driven by the introduction of textual prompts~\cite{perceiver,sd} and further refined through the use of multiple text encoders~\cite{CLIP,t5,sdxl}.

\paragraph{Latent Space.}
Early diffusion models were constrained by the computationally demanding pixel space~\cite{ddpm}. To overcome this inefficiency, newer approaches employ a two-stage process: an autoencoder first maps high-dimensional images into a lower-dimensional latent space, which is then used for the more efficient reverse diffusion process. This strategy significantly reduces computational demands, memory, and training time~\cite{sd,sdturbo}. Consequently, researchers have increasingly focused on understanding and manipulating this latent space, recognizing that working within it is key to substantially improving the quality of the output.~\cite{good_seed,hidden_language,cosed,understanding_latent}. We take advantage of this latent space, using the intermediate generations to measure the alignment of the prompt and the image being generated. This allows us to understand the generation's trajectory and predict the final alignment of the result.

\paragraph{Alignment Metrics.}
Alignment between the generated image and the text input is a key measure to assess the quality of the generative model. Dual encoders such as CLIP~\cite{clipscore,CLIP} and SigLIP~\cite{siglip} provided an early solution by directly measuring the cosine similarity between the image and text representations. More recently, approaches that leverage language models~\cite{t5} have achieved state-of-the-art results in measuring image-text similarity, demonstrating a high degree of agreement with human judgments~\cite{vqascore}. Moreover, utilizing Vision Language Models for evaluation improves generation performance by incorporating models feedback into the process~\cite{vlms_image_verification}. However, this approach requires a verifier model that may be larger than the generation model, potentially slowing the generation, and has so far only been evaluated on autoregressive models~\cite{Show-o}, excluding more common diffusion models~\cite{sd}. To mitigate this, recent work~\cite{latent_clip} employs a CLIP model to directly receive latent information from the latest diffusion step, bypassing the need to process the full image through the decoder. Additionally, another work~\cite{cfg_google} presents similar approaches to use latent information to determine the optimal classifier free guidance during the reverse diffusion process.

While current metrics successfully assess final image similarity, there's a critical need for efficient, in-process assessment during reverse diffusion. Existing latent alignment efforts either require full pipeline processing~\cite{latent_clip} or simply treat alignment as a means to an end~\cite{cfg_google}. These studies lack a dedicated analysis, open-source models, and standardized comparison data. Our work fills this gap by proposing a novel method and open dataset that leverage intermediate generation data to assess semantic alignment throughout the diffusion latent space.

\section{Aligning Language with Noisy Latents}
The core challenge of our work is to measure the alignment between the source text and the ambiguous and noisy latent representations generated during reverse diffusion. To address this, we propose a solution based on twin-tower similarity models, such as CLIP, to measure language-to-latent alignment as a similarity problem. This section starts by formalizing the language-to-latent alignment problem (Section \ref{sec:problem_definition}). We then explain how to mitigate the limited availability of noise-corrupted data (Section \ref{sec:data-generation}), required to train \model as described in Section \ref{sec:fine-tune}. Following this, we propose a new benchmark to evaluate language-to-latent assessment methods (Section \ref{sec:benchmark-proposal}).

\begin{figure*}[th]
    \centering
    \includegraphics[width=\textwidth]{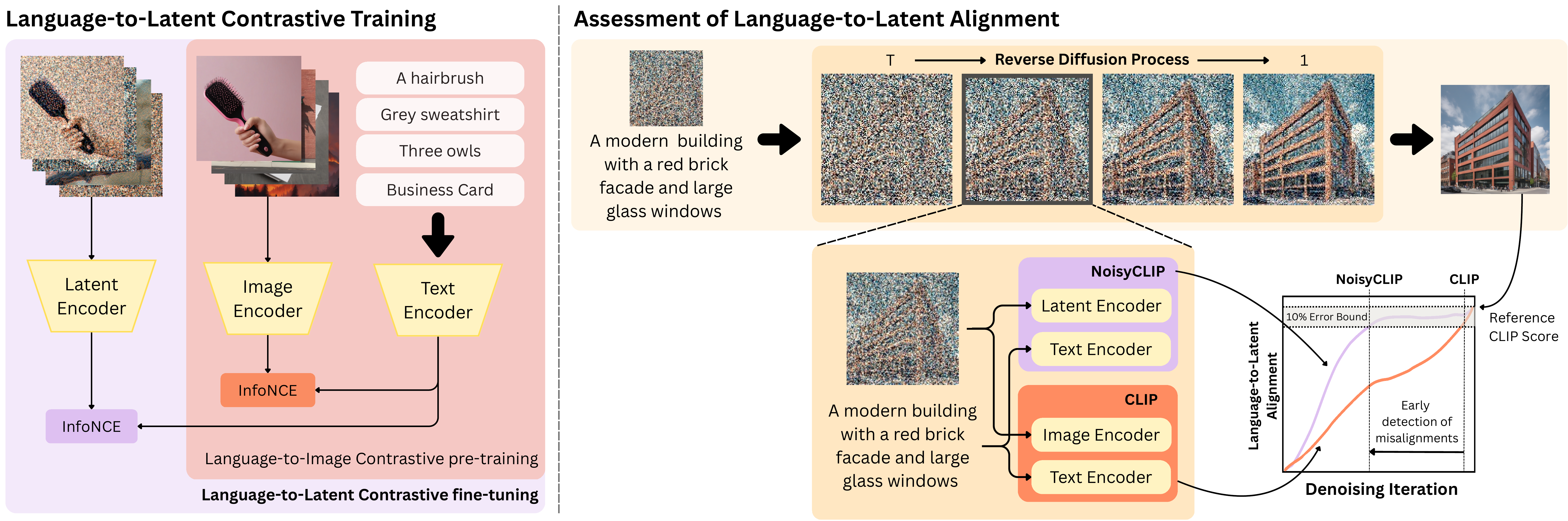}
    \caption{A prompt is used to generate an image, and at an intermediate step, the generated image and the original prompt are encoded and their similarity is measured. As demonstrated, our method generates a similarity score that is closely aligned with the score assessed for the final produced image, while at intermediate stages of generation, allowing for the early identification of misalignments.}
    \label{fig:main_figure}
\end{figure*}

\subsection{Language-to-Latent Alignment Problem}
\label{sec:problem_definition}
In image generation, earlier techniques such as Classifier-Free Guidance~\cite{cfg} enhance the output quality of diffusion models, but these approaches inherently sacrifice in-process feedback.  
This limitation leads to the adoption of post-hoc language-image alignment methods, which evaluate the final image after generation, introducing computational inefficiencies and hindering iterative control over the generation process~\cite{inference_time_scaling,vqascore}.
Despite Latent Diffusion Models~\cite{sd} grounding generation in textual prompts, there remains no mechanism to leverage this information for real-time feedback during generation.
In this section, we explore how to enable direct language-to-latent alignment evaluation during generation, eliminating post-hoc assessment and offering a more scalable and responsive approach to image synthesis. We begin by recalling the objective function of the Latent Diffusion Model:
\begin{equation}
L_{LDM} := \mathbb{E}_{\mathcal{E}(x), y, \epsilon \sim \mathcal{N}(0, 1), t} \left[ \left\| \epsilon - \epsilon_{\theta}(z_t, t, \tau_{\theta}(y)) \right\|_2^2 \right]
\end{equation}
In this formulation, $x$ is the original image, and $z=\mathcal{E}(x)$ is its latent representation generated by the encoder $\mathcal{E}$. $z_t$ represents the noisy latent state after $t$ diffusion steps, where $t$ ranges from $1$ to $T$. $\epsilon$ is the Gaussian noise used to corrupt the latent state, and $\epsilon_{\theta}$ is the noise prediction model trained to estimate the noise $\epsilon$ added to $z_t$. The reverse diffusion steps are conditioned by the input text $y$, encoded by the text encoder $\tau_{\theta}(y)$. The text embedding $\tau_{\theta}(y)$ is then used in cross-attention mechanisms to interact with noisy latent states $z_t$, guiding the denoising process towards a text-aligned image. We maintain this standard notation throughout the paper.

To evaluate similarity in the image space, we calculate the cosine similarity between the image and text prompt~\cite{CLIP}. 
Given the final generated image, $x$, and the text prompt, $y$, we employ the text encoder, $\tau_{\theta}$, and an image encoder, $\nu_{\theta}$, to compute the similarity score,

\begin{equation}
S_{\text{final}}({x}, {y}) = \tau_{\theta}({y})\cdot\nu_{\theta}({x}).
\end{equation}

To generalize this operator to intermediate diffusion steps, we extend $S_{\text{final}}({x}, {y})$ which operates on  pixel-space images ($x$), to support latent representations $z_t$ that cannot be directly inputted into $\nu_{\theta}$. This extension is defined as

\begin{equation}
S_{\text{latent}}({z}_t, {y}) = \tau_{\theta}({y})\cdot\nu_{\theta}(\Phi({z}_t)),
\end{equation}
where $z_t$ is the latent representation at step $t$, and $\Phi$ is a transformation (inverse of $\mathcal{E}$) that converts $z_t$ into an RGB-like space consumable by the image encoder $\nu_{\theta}$, as illustrated in Figure~\ref{fig:main_figure}. The text prompt $y$ and the text encoder parameters remain the same as in $S_{\text{final}}$.

The definition of $S_{\text{latent}}$ allows for the assessment of how well image representations align with text prompts during the reverse diffusion process. This approach relies on having access to noise-corrupted data. In the following section, we will explore the scarcity of publicly available data of this kind and explore our approach to generate this kind of data.

\subsection{Sampling Noise-Corrupted Data}
\label{sec:data-generation}

There is a notable lack of open data involving noisy latent images, as prior works often rely on on-the-fly noise inversion~\cite{object_inversion,prompt_tuning_inversion,dire}. This hinders reproducibility and consistent benchmarking. To mitigate this, our training dataset $\mathcal{D}$ is structured as a collection of samples where each sample within $\mathcal{D}$ is a tuple of the form $(y, x, \{z_t\}_{t=1}^T)$, where $y$ represents the image generation prompt, $x$ is the corresponding generated image and $\{z_t\}_{t=1}^{T}$ represents the sequence of latent representations sampled from the image's reverse diffusion process.

We optimize our encoder using a rewritten version of CC12M~\cite{liu2024llavanext,conceptual-captions-cc12m-llavanext}. These refined captions provide dense, detailed prompts that better align with the requirements of image generation tasks, based on the diverse visual concepts present in the original CC12M~\cite{cc12m}. Using SDXL~\cite{sdxl}, we generate images from these prompts and extract latent representations for training (as detailed in Section~\ref{sec:implementation}). 

Although this process successfully captures the complete latents necessary for optimization, we cannot simply sample latents at random throughout the sequence. Early steps are dominated by Gaussian noise, whereas late steps lack the ambiguity required to train an early-stopping monitor. Determining the optimal noise distribution therefore becomes a highly non-trivial challenge, motivating our systematic evaluation in Section~\ref{sec:impact_latent_training}.

\subsection{Language-to-Latent Contrastive Training}
\label{sec:fine-tune}
To improve noise robustness in pre-trained vision encoders, we introduce \model, a twin-tower encoder capable of measuring similarity earlier in the reverse diffusion process.

We select a fixed latent interval (20–29) for contrastive training to enable stable language-to-latent alignment across diverse representations using only a subset of the data, as validated in Section~\ref{sec:impact_latent_training}. Using the generated noisy latents $z_t$, we exclusively fine-tune the image encoder (see Figure~\ref{fig:main_figure}). During the contrastive learning, each noisy latent representation $z_t$ forms a positive pair with its corresponding text prompt $y$, while being contrasted against in-batch negatives $\mathcal{N}$ (all image-text pairs in the batch except the positive pair).

By applying the contrastive loss (InfoNCE)~\cite{infonce} we maximize similarity between positive pairs $(z_t,y)$ while suppressing similarity with negatives $\mathcal{N}$.
Freezing the text encoder maintains its general-purpose functionality while optimizing language-to-latent alignment through image encoder tuning. The crucial advantage of this approach is that the frozen text encoder can be shared with the diffusion model, which ensures consistency between textual representations and the diffusion model's conditioning space. This grounding forces the latent encoder to align better with the diffusion representations, effectively mimicking the cross-attention interaction during denoising while providing an early projection of final image alignment~\cite{perceiver}.

\subsection{Benchmark Proposal}
\label{sec:benchmark-proposal}
Lastly, to overcome the lack of standardized data to evaluate representations in inner diffusion processes, we propose a benchmark built on top of image captioning~\cite{cc12m} and image generation~\cite{genai-bench} datasets. We derive two subsets, Noisy-Conceptual-Captions and Noisy-GenAI-Bench to enable open, systematic community evaluation across three critical dimensions.

\paragraph{Diverse visual representations.}
To evaluate inner diffusion representations, we generate multiple images for each prompt $p$, allowing us to quantify how alignment metrics choose the output that most closely matches the prompt across diverse visual representations. 

\paragraph{Factual consistency.} We employ a classification task where a single image is evaluated against $k$ prompts, one correct and $k-1$ intentionally corrupted to systematically violate factual consistency (as detailed in the Appendix). This approach improves the robustness of the evaluation compared to random prompt sampling by introducing targeted factual violations rather than random information. 

\paragraph{Noisy latent information.}
To support the assessment of the previously described tasks in the reverse diffusion process, we store the noisy latent information at each denoising step during generation.

\section{Experimental Setup}
In this section, we provide a detailed description of our evaluation settings, including tasks, datasets, baseline models, metrics, and implementation details.

\subsection{Evaluation Tasks}
To evaluate the proposed method, we conducted the following experiments:

\begin{itemize}
    
    \item{\textbf{Language-to-Latent Alignment.}} We start by examining the language-to-latent alignment throughout the denoising process of N competing random seeds. We measure the alignment with \model and CLIP, and compare the difference between the best and the worst images (VQAScore is used to rank images by their alignment with the prompt).
    This allows us to observe how early \model can detect misalignments.
    \item{\textbf{Latent-to-Text Retrieval:}} As an additional experiment, we assess the model's ability to assign the highest score to the most factually accurate prompt. This experiment aims to analyze \model in the presence of captions that include non-factual distractors. This evaluation is crucial as it tests the model's capacity to distinguish the correct caption from highly similar, yet non-factual, errors. Ultimately, success on this task suggests that the model possesses the semantic understanding required to prevent non-factual image selections during generation.
    \item{\textbf{Early-Stopping Best-of-N:}}
    Our main experiment follows recent work~\cite{vqascore, inference_time_scaling,xie2025sana} that demonstrated a significant improvement in image quality by generating multiple images and selecting the most aligned one. 
    In our setting, instead of completing all denoising processes, we leverage \model early misalignment detection capabilities, and evaluate how early it can stop poorly aligned denoising processes and keep the best language-to-latent aligned one. 
\end{itemize}

\subsection{Datasets}
For the experiments explained above, we use two datasets. Specifically, the test sets are:
\begin{itemize}
    \item \textbf{Noisy-Conceptual-Captions}: This subset from our benchmark comprises a set of 1000 captions sampled to enable objective assessment of factual caption selection. This dataset comprises the original prompts, four generated images per original prompt as well as four non-factual prompts (the creation of these prompts is detailed in the Appendix).

    \item \textbf{Noisy-GenAI-Bench}: This dataset includes a total of 1600 prompts, consisting of 730 basic prompts and 870 more advanced ones, going from simple compositions such as objects (e.g. "a dog") to more complex compositions such as counting or negations. To evaluate the performance on the BoN task, we expand this dataset by generating 10 images along with the corresponding latent information for each prompt.
\end{itemize}

\subsection{Baselines}

We evaluate \model against several established baselines. Because our approach builds upon CLIP~\cite{CLIP}, we utilize pre-trained CLIP and SigLIP~\cite{siglip} as primary points of comparison, employing their base versions for ablations and large versions for main results. We also compare against LatentCLIP~\cite{latent_clip} (versions 8 and 4-plus) to benchmark against latent-optimized representations. All primary generation experiments are conducted using Stable Diffusion XL~\cite{sdxl}, with generalization tests performed on two DiT-based~\cite{dit} architectures: Stable Diffusion 3.5~\cite{sd3} and FLUX.1 [dev]~\cite{labs2025flux}.

\subsection{Evaluation Metrics}

In the Latent-to-Text Retrieval task, we presented each image with a single correct prompt and multiple non-factual prompts, measuring performance using Recall@1 (R@1)~\cite{longclip,seeing_what_matters,distinctive_captioning,waffle}. This metric quantifies the model's accuracy by measuring the proportion of images in which the correct prompt is ranked first among all candidates.

For the BoN task, we assess the model’s ability to select the most relevant image for a given prompt. VQA-based evaluation methods have shown improved reliability and alignment with human assessments~\cite{visual_programming_t2i,tifa,t2i_compbench,genai-bench,what_you_see}. Similarly to previous research, we adopt VQAScore~\cite{vqascore} as our primary metric to measure how well encoders can select the most similar image over others for a certain prompt~\cite{gender_biases_t2i,generalizing_alignment,nl_inference_vision_language,decoder_llm_t2i,imagen3}.

\subsection{Implementation Details}
\label{sec:implementation}

\textbf{Training.} We fine-tune solely the visual encoder of the dual-encoder on a dataset of 50000 latents, which are uniformly sampled from the generated images. Training is done with a learning rate of 5.7e-5, using a cosine annealing scheduler with a warm-up ratio of 0.1 and a weight decay of 0.1. All computations are performed on an NVIDIA A100 40G GPU over 10 epochs with a batch size of 128. \textbf{Latent RGB Representation.} Although SDXL~\cite{sdxl} latent space utilizes a four-channel structure, direct visualization requires transforming these latents into RGB space. This transformation is done by decoding latent representations via a linear transformation to RGB~\cite{sdxl_latent}. For our generalization experiments on DiTs, the highly compressed nature of their latent spaces requires using the model's VAE for RGB decoding.

\section{Results and Discussion}

In this section, we evaluate how $\text{\model}$ addresses the challenges established in our introduction. We begin by assessing $\text{\model}$'s Language-to-Latent Alignment to validate its capacity for monitoring intermediate denoising steps. To understand the data strategies required for this monitoring, we then analyze the Impact of Training Latent Ranges on Inference. Following this, we demonstrate practical computational reductions by employing our method for Early-Stopping in Best-of-N Sampling. Finally, we present a Generalization Analysis across varying backbone architectures (further ablations in the Appendix).

\subsection{Language-to-Latent Alignment}
\label{sec:score_comparison}

\subsubsection{Language-to-Latent Alignment Estimators.}
To evaluate alignment effectiveness, we sample prompts from the Noisy-Conceptual-Captions dataset, generating four images per prompt. Using VQAScore to establish a reference ranking, we report the average similarity score assigned to each rank by both~\model and CLIP.

\begin{figure}[t]
     \centering
     \begin{subfigure}[b]{0.32\textwidth}
         \centering
         \includegraphics[width=\textwidth]{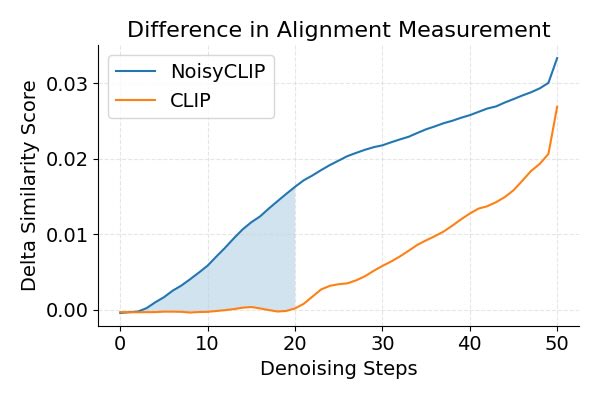}
     \end{subfigure}
     \hfill
     \begin{subfigure}[b]{0.32\textwidth}
         \centering
         \includegraphics[width=\textwidth]{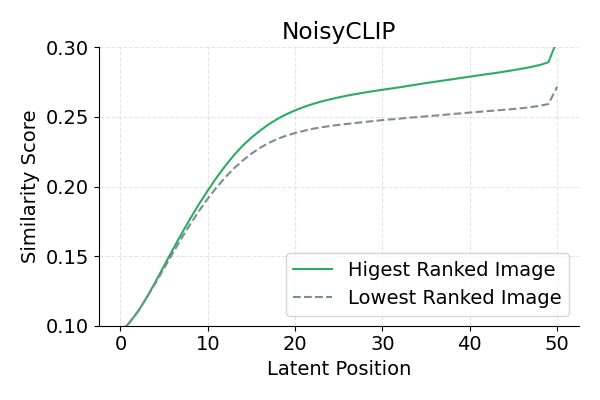}
     \end{subfigure}
     \hfill
     \begin{subfigure}[b]{0.32\textwidth}
         \centering
         \includegraphics[width=\textwidth]{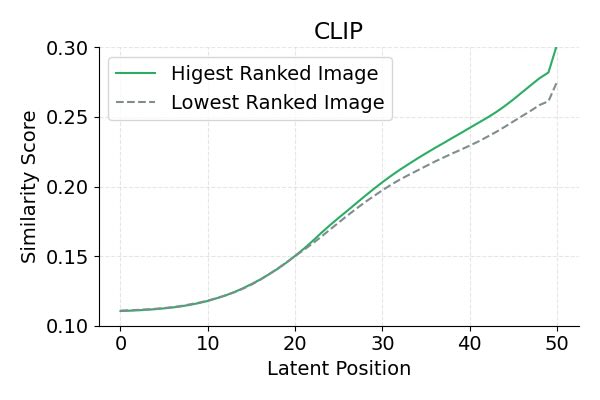}
     \end{subfigure}
     \caption{Analysis of alignment measurement throughout the reverse diffusion process shows that our model distinctly separates aligned from misaligned images as soon as latent 20.}
     \label{fig:our_approach_scores}
\end{figure}

As illustrated in Figure~\ref{fig:our_approach_scores}, our model effectively discriminates the top image from sub-optimal ones as early as latent 20, as the backbone can only do so by latent 40. We can see that our method increases the delta between the highest and lowest scoring images surpassing 2\% in the middle of the generation. Although our approach focuses on the noisy images, the results do not show any deterioration in the image dimensions. Furthermore, our approach achieves even greater differentiation by the end of the process compared to the backbone trained on images alone.

In contrast, CLIP exhibited persistent difficulty maintaining a separation between the highest and lowest scoring images. Although it reaches a 2.5\% delta in the image dimension, the gap remains under 1\% for most of the generation (up to the mid-30s) and only emerges after latent 20. These results demonstrate that our approach reliably measures similarity in intermediate latents while achieving final scores comparable to frozen CLIP. Moreover, our model produces scores that increase almost linearly from the start of generation, whereas CLIP only shows this behavior from the midpoint onward.

\subsubsection{Latent-to-Text Retrieval.}

\begin{wrapfigure}[13]{l}{0.45\textwidth}
    \centering
    \vspace{-15pt}
    \includegraphics[width=\linewidth,trim={0cm 0cm 0cm 0cm},clip]{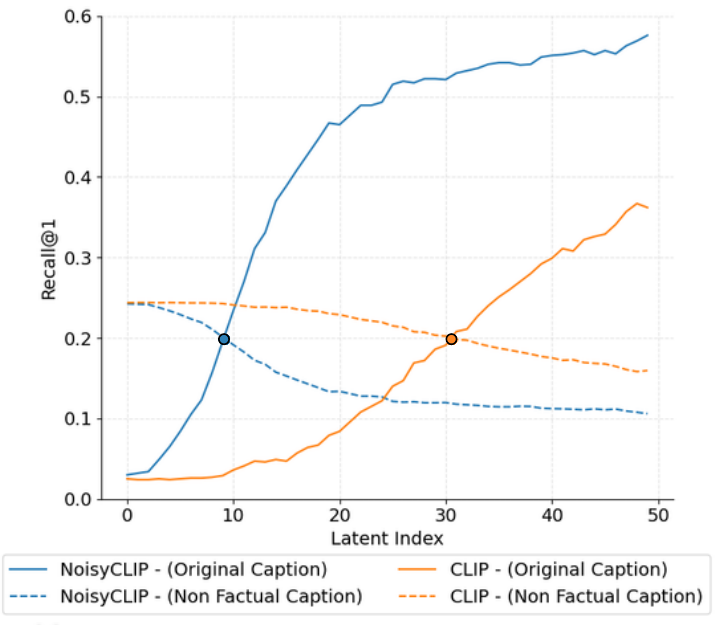}
    \vspace{-17pt}
    \caption{Recall@1 for Caption Selection.
    }
    \label{fig:recall_i2t}
\end{wrapfigure}

This experiment evaluates models ability to select the correct caption for an image from five candidates (one correct, four non-factual). As shown in Figure~\ref{fig:recall_i2t}, \model discriminates the correct caption from the non-factual ones remarkably early, by latent 10. In contrast, CLIP requires an additional 20 denoising iterations to achieve a similar separation (colored dots, Figure~\ref{fig:recall_i2t}).

While both models increasingly favor the correct caption as denoising progresses, \model demonstrates significantly faster and higher convergence. Within the first 25 iterations, \model's recall@1 for the correct caption surges from near zero to over 0.5, whereas the baseline remains below 0.1. Furthermore, \model's recall stabilizes during the second half of generation, whereas the baseline never exceeds 0.4 (further analysis of non-factual caption selection is provided in the Appendix).

\subsection{Impact of Training Latent Ranges on Inference}
\label{sec:impact_latent_training}

\begin{wrapfigure}[13]{r}{0.5\textwidth}
    \centering
    \vspace{-20pt}
    \includegraphics[width=\linewidth,trim={0cm 0.1cm 0cm 2cm},clip]{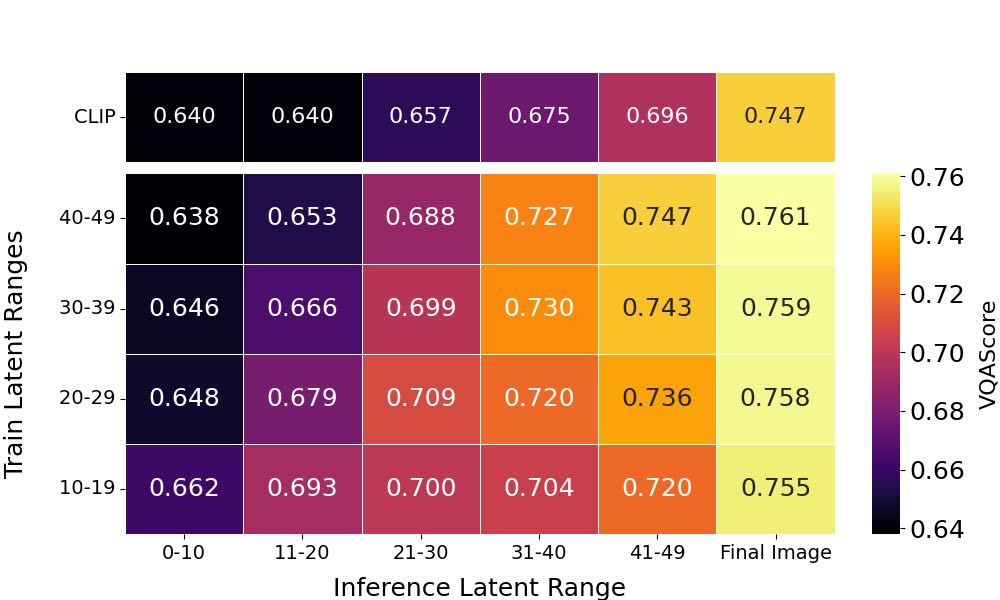}
    \caption{Performance across inference ranges (columns) when trained with different latent ranges (rows).}
    \label{fig:latent_evaluation}
\end{wrapfigure}
In this experiment we examine how different training latent ranges impact \model performance.
Results are reported on the Noisy-Concept-Captions dataset. 
As shown in Figure~\ref{fig:latent_evaluation}, the alignment performance correlate with the inference latent range: lower latent ranges result in less accurate alignment, underscoring the challenge of assessing language-to-latent alignment under the presence of noise.

The baseline CLIP model (top row), performs poorly in latents ranges 1-49 achieving a top performance of 0.696, despite a VQAScore of 0.747 on the final image. This disparity highlights the limitation of lacking noisy data training. By comparison, \model inference performance generally correlates with the level of noise present during training. Low-noise training (ranges 40-49) consistently yields the best overall results in both the final latent range and the final image. However, the model trained on initial latent ranges (10-19) performs best early on (up to 4 points gain) but declines sharply in the final two ranges. Middle-range training (20-29) strikes the best balance, achieving stable performance: it is comparable to the best models (40-49) in later inference ranges while performing better in initial ranges. These results indicate that moderate training noise (20–29) optimizes balanced performance across inference stages.

\subsection{Early-Stopping in Best-of-N Sampling}
\label{sec:best_of_n}

To further evaluate our fine-tuning approach with noisy latent information, we analyze~\model performance as a ranker on a BoN denoising task. We evaluate model’s ability to select a candidate image by denoising multiple seeds with the same prompt, and scoring the latents according to its alignment with the prompt in the GenAI-Bench basic prompts. This enables the early stopping of denoising processes that are estimated to be sub-optimal, reducing the aggregated number of denoising iterations. A detailed analysis of the performance on the basic and advanced prompts is provided in the Appendix.

\subsubsection{Defining Denoising Iterations as a Metric for Cost.}
We define cost by the number of reverse diffusion steps required to produce an image; in our baseline~\cite{sdxl}, this is 50 steps per image. If BoN is applied at a certain stage, only one image proceeds with the generation. For example, if two images begin generation but only one is selected to continue at latent stage 25, the total cost becomes $(25 \times 2) + (25 \times 1) = 75$, representing a 25\% saving compared to a full "best-of-2" approach ($2 \times 50 = 100$).

\begin{wrapfigure}[12]{r}{0.5\textwidth}
    \centering
    \vspace{-20pt}
    \includegraphics[width=0.9\linewidth]{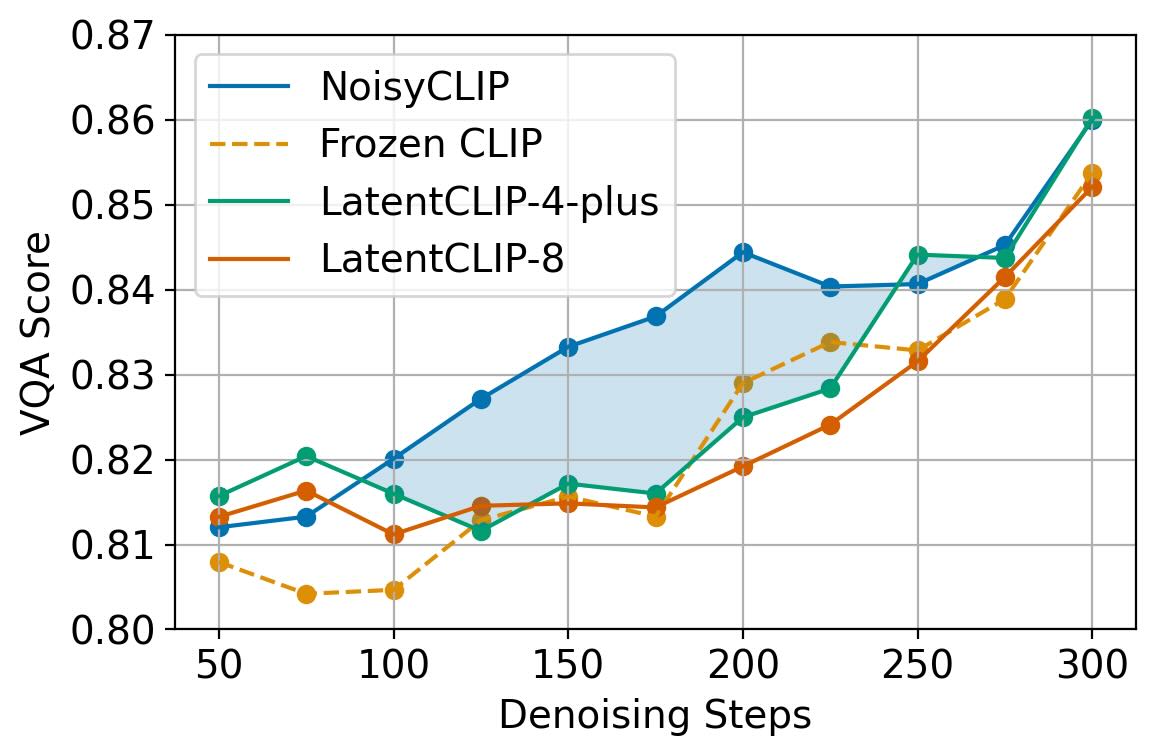}
    \vspace{-10pt}
    \caption{Alignment (VQAScore) vs Best-of-N denoising cost. Cost is normalized for a scenario where $N=6$.}
    \label{fig:cost_alignment}
\end{wrapfigure}
\subsubsection{Performance Gains Versus Costs Reduction.}
\label{sec:bon-sdxl}
In Figure \ref{fig:cost_alignment} we analyze the trade-off between prompt alignment and computational cost within a specific best-of-6 selection scenario. The standard baseline, which uses CLIP to select the final image after all 6 generations are complete, achieves a VQAScore of 0.85 at a cost of 300 ($=6\times50$). At this same 300-step cost, the best result achieved by LatentCLIP matches the performance of our method (0.86).

By reducing the computational budget by 50\% (to 150 denoising steps), our method incurs a minimal 2\% drop in VQAScore compared to 300-step CLIP and LatentCLIP-8 baselines, and less than a 3\% drop compared to LatentCLIP-4-plus. Furthermore, when evaluated at this reduced 150-step budget, our approach outperforms all baselines by nearly 2\%. This advantage persists at a 200-step budget (a one-third reduction), where our method continues to achieve the highest latent selection score. Overall, our approach consistently outperforms the baselines across the reverse diffusion process, demonstrating a superior balance between alignment and efficiency.

\subsubsection{Selecting images in the latent space.}

\begin{wrapfigure}[15]{l}{0.55\textwidth}
    \centering
    \vspace{-25pt}
    \includegraphics[width=0.85\linewidth]{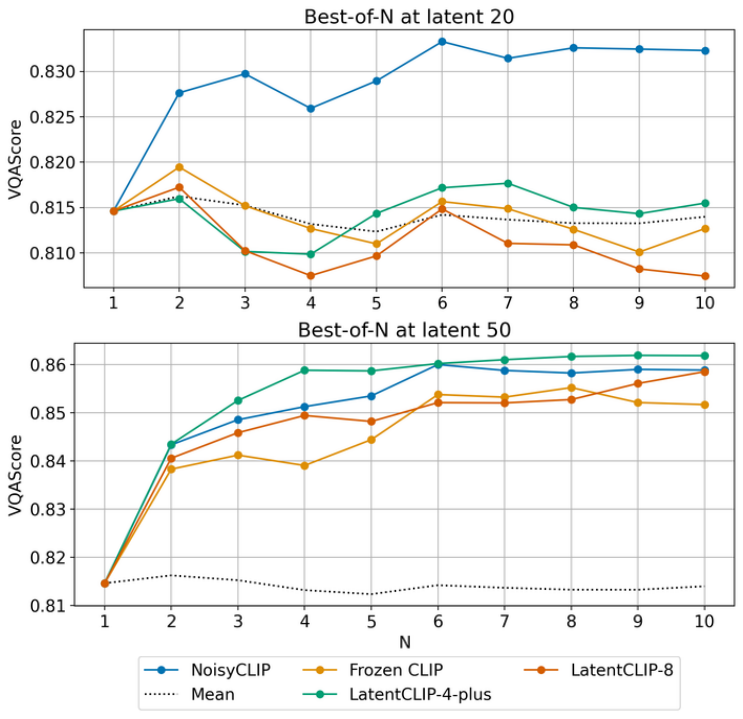}
    \vspace{-10pt}
    \caption{Best-of-N results for latent 20 and final image. Alignment is assessed using VQAScore.}
    \label{fig:bon6}
\end{wrapfigure}

In this analysis, we compare the performance of our method on Best-of-N denoising task, with $N=\{1,\ldots,10\}$.
When applying a dual encoder to rank the N generations in latent 20 (Figure~\ref{fig:bon6}, top), the perfomance of all baselines mirror the mean alignment of the entire candidate set. This suggests that applying current approaches at this stage yields negligible benefit. In contrast, our method improves alignment by up to 2\%, demonstrating the effectiveness of our noise-aware approach in handling early-stage latent representations.

Figure~\ref{fig:bon6} (bottom) presents results for the compute-intensive final image selection, where all images are fully generated. While CLIP baseline benefits from the absence of noise in the final images, our method consistently achieves superior prompt alignment across all values of $N$. Even compared to LatentCLIP (trained on clean $z_0$ latents), our model consistently outperforms their version with a patch size of 8 and remains highly competitive with the 4-plus variant.

In summary, \model enables effective mid-generation BoN selection, while baselines fail in high-noise settings by mirroring random sampling. Though competitive with LatentCLIP in final stages, our novelty lies in unlocking computational efficiency via early-step ranking, where current CLIP methods fail.

\begin{figure}[t]
    \centering
    \includegraphics[width=\textwidth]{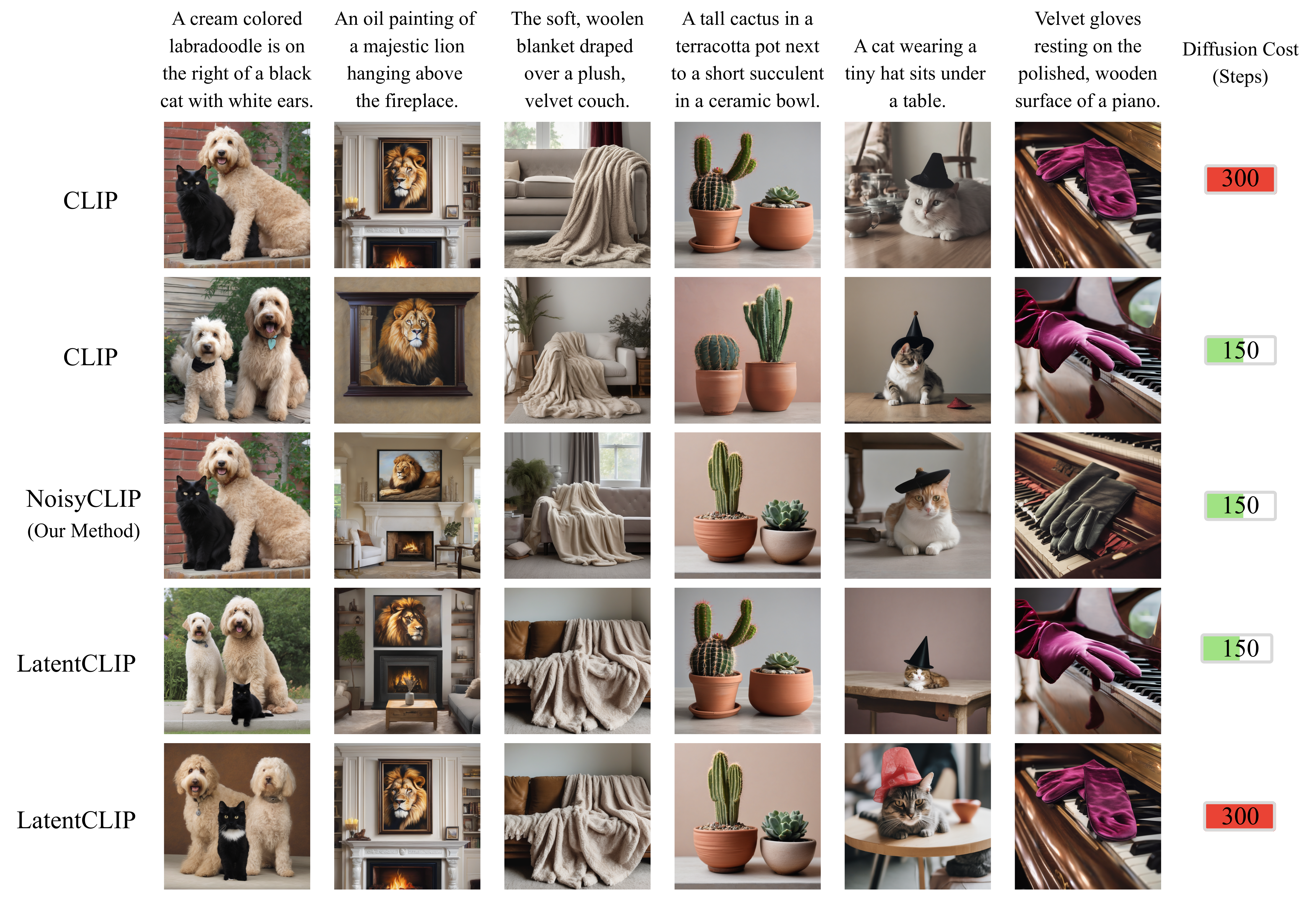}
    \caption{Images generated using a best-of-6 strategy. We compare \model with the CLIP and LatentCLIP-4-plus under two computation budgets: 150 and 300 diffusion steps (generating all 6 images). Our approach surpasses the baseline at the 150-step limit and achieves comparable or superior alignment to the full 300-step baselines.}
    \label{fig:qualitative_samples}
\end{figure}

\subsubsection{Qualitative Results.}
Figure~\ref{fig:qualitative_samples} demonstrates that our method achieves high-fidelity image selection in a best-of-6 setting using only 150 diffusion steps, consistently matching or surpassing CLIP and LatentCLIP at an equivalent or higher computational cost. For simple queries (e.g., the third prompt), all approaches successfully select text-aligned images, confirming that standard generative models adequately handle low-complexity prompts. 

However, our method excels on complex compositions. For \textit{"a tall cactus in a terracotta pot next to a short succulent in a ceramic bowl"}, only \model maintains all attributes (plant heights, container materials) using half the computation. Similarly, \model resolves subtle relationships, identifying a cat under a table (fifth prompt) that baselines miss at full computation. Furthermore, it distinguishes resting gloves on a piano (sixth prompt) at half the budget, a nuance where baselines fail completely. By capturing these details early in the latent space, our results validate that alignment can be measured without the overhead of full image generation (see Appendix for failure cases).

\begin{wrapfigure}[11]{r}{0.5\textwidth}
    \vspace{-60pt}
    \centering
    \includegraphics[width=0.9\linewidth]{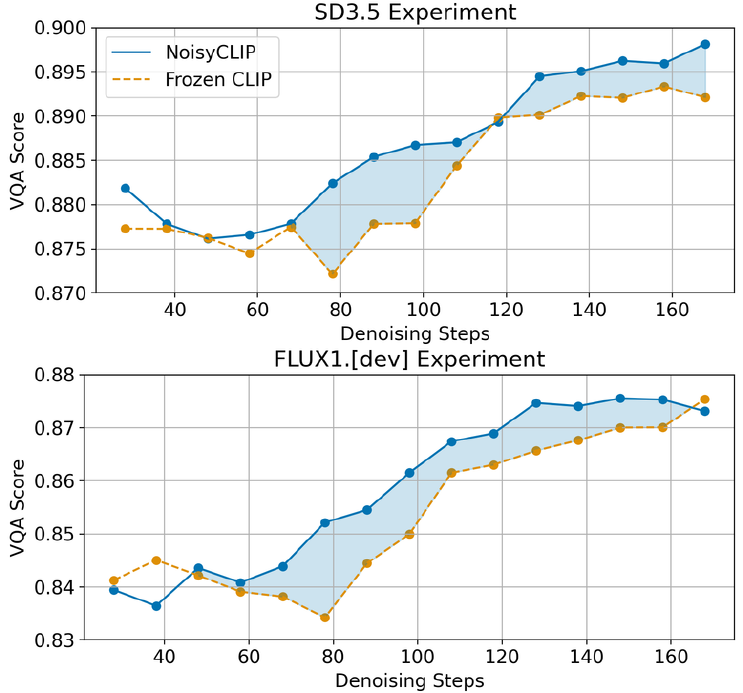}
    \vspace{-10pt}
    \caption{Alignment versus computational cost for zero-shot generalization on DiTs.}
    \label{fig:generalization}
\end{wrapfigure}

\subsection{Generalization Analysis}

\subsubsection{Diffusion Architectures.}

To evaluate the robustness of NoisyCLIP, we validate its generalization capabilities on two DiT architectures in a zero-shot setting. By applying NoisyCLIP to VAE-decoded latent representations, we observe consistent improvements over the frozen CLIP baseline across these distinct models. Specifically, NoisyCLIP achieves a +2\% alignment gain on Stable Diffusion 3.5~\cite{sd3}  and a +1.8\% gain on FLUX.1~\cite{labs2025flux}, alongside a 57\% reduction in denoising steps (Fig.\ref{fig:generalization}).

These findings reinforce two core strengths of our method. First, they demonstrate that although NoisyCLIP is trained exclusively on data generated by SDXL, our noise-aware contrastive training effectively transfers to different generative architectures. Second, the observed alignment gains mirror those achieved on SDXL (Sec.~\ref{sec:bon-sdxl}), demonstrating that our significant reductions in computational cost persist even when applied to newer models configured with fewer denoising iterations (e.g., 28 steps).

\subsubsection{NoisyCLIP Backbones.}

\begin{wraptable}[13]{r}{0.5\textwidth}
    \centering
    \small
    \vspace{-35pt}
    \caption{Performance comparison in latent and image spaces. We report Factual
Consistency (R@1) and Best-of-N (VQAScore).}
    \label{tab:overall-results}
    \setlength{\tabcolsep}{3pt}
    \begin{tabular}{@{}lcc@{}}
    \toprule
    \textbf{Model} & \textbf{Fact. Consist.} & \textbf{BoN} \\ 
    \midrule
    \multicolumn{3}{l}{\textit{Denoised Latent (it. 21-30)}} \\
    \model & \textbf{0.506} & \underline{0.709} \\
    CLIP L14 & 0.145 & 0.657 \\
    \modelsiglip & \underline{0.194} & \textbf{0.718} \\
    SigLIP L16 & 0.147 & 0.652 \\ 
    \midrule
    \multicolumn{3}{l}{\textit{Final Image (it. 50)}} \\
    CLIP L14 & -- & 0.747 \\
    SigLIP L16 & -- & 0.764 \\ \bottomrule
    \end{tabular}
\end{wraptable}

Table~\ref{tab:overall-results} shows our approach substantially improves intermediate alignment. In factual consistency (R@1), \model boosts CLIP performance by $3.5\times$. On the Best-of-$N$ task, it achieves a 0.709 VQAScore (+5\% over baseline), reaching 95\% of final-image performance midway through generation. 

Evaluating backbone transferability, \modelsiglip matches \model on our primary Best-of-$N$ visual ranking task (0.718 vs. 0.709). Although \modelsiglip trails in R@1, this discrepancy is an expected architectural consequence. R@1 tests fine-grained text perturbations, and SigLIP's sigmoid loss inherently lacks the discriminative sharpness of CLIP's softmax for subtle textual nuances~\cite{beta_clip}. Thus, our method transfers robustly across backbones for visual selection, with text-discrimination performance constrained by the base encoder.

\section{Conclusions}
In this work, we introduced \model, a  framework for evaluating language-to-latent alignment during the reverse diffusion process by quantifying similarity across intermediate latent representations.
Empirical evaluations demonstrate that \model provides substantial advantages in Best-of-N generation strategies by enabling early-stage discrimination of higher-quality samples. This capability leads to more efficient use of computational resources, reducing the number of required diffusion iterations by up to 50\% while preserving the fidelity of the final outputs relative to a full cost solution.

Collectively, these findings establish the feasibility of continuous, in-process alignment evaluation within latent diffusion models. By leveraging iterative latent information, \model transforms alignment assessment from a purely post-hoc procedure into an integral component of the generation process advancing the development of adaptive, controllable, and computationally efficient multimodal generative systems.

\clearpage
\noindent\textbf{Acknowledgments.} We thank anonymous reviewers for their valuable comments and suggestions. This work was partially supported by the FCT Ph.D. scholarship grant Ref. 2025.02103.BD, by the FCT project NOVA LINCS Ref. (UID/04516/2025) and by a Google Research Gift.

\bibliographystyle{splncs04}
\bibliography{main_nourl}

\clearpage
\appendix

\section{Non factual image-text pairs generation}
\label{a:non-factual-pairs}

In this section, we explain how we generate the necessary prompt corruptions for the Factual Consistency task, where encoders are employed to choose the most accurate prompt for a given image.
Avoiding random LLM corruption, we define four specific dimensions of non-factuality to be corrupted (defined as ERROR\_TYPE in the LLM prompt)~\cite{hallucination_snowballing,visual_hallucnations_mmllm,MMIR}. 

The dimensions are as follows: 
\begin{enumerate} 
    \item \textit{Color}, which involves changing the color of the primary subject (e.g., an original caption with "a red apple" becomes "a green apple" in the corrupted version).
    
    \item \textit{Count}, which involves changing the number of the principal subject (e.g., if the original caption says "two dogs," the corrupted version changes it to "three dogs"). 
    
    \item \textit{Background}, which involves changing the setting of the main subject (e.g., a  caption states "a cat on a mat" transforms to "a cat on a table" in the corrupted version). 
    
    \item \textit{Main subject}, which involves changing the principal subject itself (e.g., original caption says "a bird on a branch," the corrupted version becomes "a squirrel on a branch").\\ 
\end{enumerate} 
To generate the misalignments, we use the instruction-tuned version of Gemma 3 with 27B parameters~\cite{gemma-3}. We employ this model to rewrite each original prompt based on the defined dimensions using the following prompt:
\begin{spverbatim}
You are a highly specialized text transformation AI. Your sole task is to receive an original sentence and generate a single, corrupted version of that sentence.

Rules for Corruption:
1.  Maintain Original Structure:  The overall sentence flow must remain identical to the original.

2.  Targeted Non-Factuality: Only modify the aspect of the sentence corresponding to the '{ERROR_TYPE}' provided. This modification should render that specific aspect non-factual or illogical within the context of the original sentence.

3.  Single Output: Output only the corrupted sentence. No additional text, explanations, or formatting.

4. Coherent Output: Ensure that the output is coherent with the new changes.

Change the {ERROR_TYPE} in the following PROMPT:
PROMPT: {PROMPT}
\end{spverbatim}

\section{Selecting the correct prompt}

\begin{wrapfigure}[13]{r}{0.5\linewidth}
    \centering
    \vspace{-20pt}
    \includegraphics[width=\linewidth]{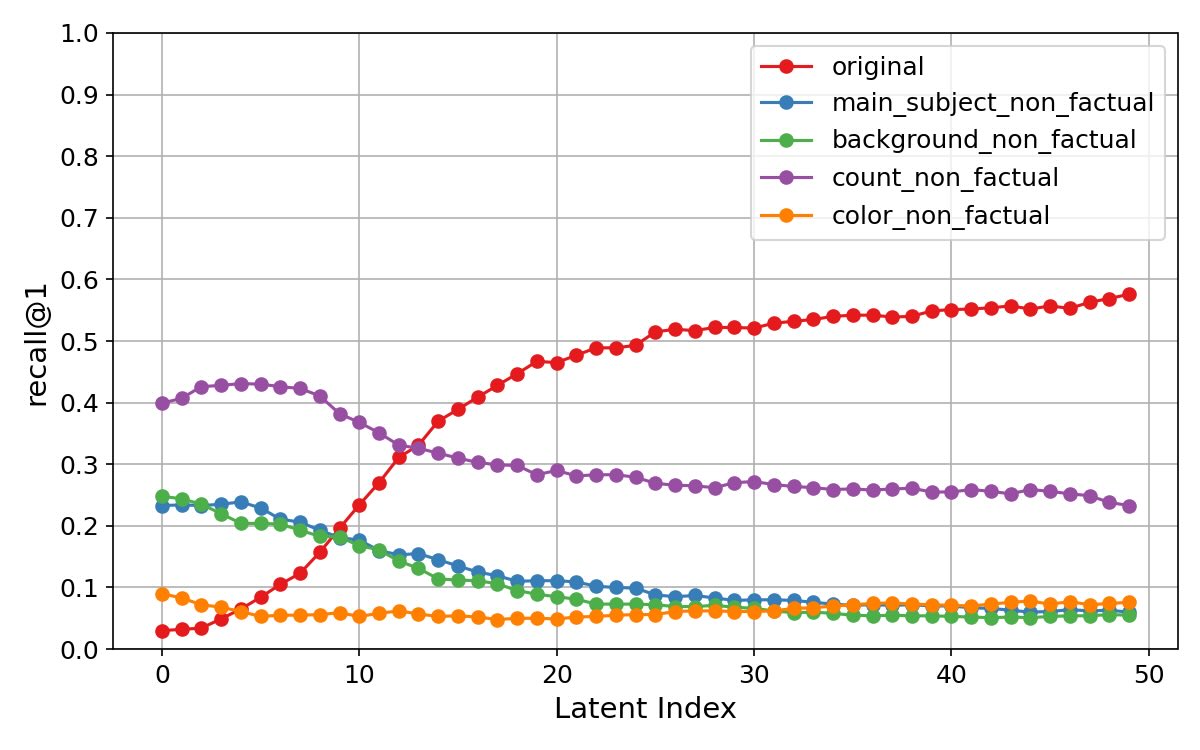}
    \vspace{-20pt}
    \caption{Recall@1 in Caption Selection. Compares different prompt types, showing how different prompt corruptions affects the model's caption selection.}
    \label{fig:correct_prompt}
\end{wrapfigure}

Furthering our analysis on Factual Consistency we now discuss how our method distinguishes different non-factual prompts and how that can influence the final results.

Analyzing recall@1 in Fig.~\ref{fig:correct_prompt}, we compare how~\model selects the prompts along the generation. An overview shows that our model clearly distinguishes the accurate caption in the early latents. Despite the initial noise, our model reaches a recall@1 value greater than 0.5 in the middle of the generation.

As generation progresses, the selection of non-factual captions declines. The 'count' caption consistently maintains the highest recall among the non-factual options, this selection of a wrong captions is likely due to CLIP's known limitations in counting due to its training on single-object images \cite{clip_counting_cvpr,clip_counting_cvpr2022,clip_counting_iccv}. Considering the remaining non-factual captions from the middle of generation onward, the recall remains below 0.1, reinforcing the effectiveness of our approach when evaluating highly similar captions. Notably, the 'color' non-factual caption remains consistently below this threshold, while non-factual changes related to artifacts (subject or background) show similar selection behavior.

This analysis concludes that our approach's performance in ranking the non-factual alternatives is heavily influenced by input type. Specifically, we observed that "count" prompts yielded the highest recall, whereas "color" prompts showed the lowest, reinforcing known biases inherent to CLIP models.

\section{Deeper analysis on  Best-of-N Results}

This analysis provides a more in depth analysis of twin-tower encoders selection in Noisy-GenAI-Bench using a best-of-10 evaluation approach. During this section, we assess the model’s capabilities across ten skill dimensions, split into basic and advanced prompts.

GenAI-Bench defines basic skills as core abilities such as, identifying visual attributes (e.g. color, material), contextual backgrounds, spatial relationships (e.g. an entity facing another), action dynamics (e.g. an entity pushing another), and part-whole configurations (e.g. a component is part of another, such as a body part). Simultaneously GenAI-Bench defines a set of advanced skills too, including counting, differentiation within categories (e.g. distinguish between two objects of the same category where one is "young" and another is "old"), comparative characteristics (e.g. in a classroom there are more people standing than sitting) , negation of attributes, and universal applicability (e.g. in a parking lot all cars are red). 

We test selections of ~\model, CLIP, LatentCLIP-4-Plus and LatentCLIP-8 during generation by choosing the best image between latents 21 and 30. Since CLIP was trained on non-noisy images and LatentCLIP on the last latent iteration, we also apply them to the final images. For this experiment, we report the VQAScore to show how well the selected images align with the prompts.

\begin{table}[h]
\centering
\caption{Results of applying the test-time scaling with Best-of-N during the generation process on the basic prompts of the GenAI-Bench.}
\label{tab:basic_genai_bench}
\begin{tabular}{lccccc}
\hline
\multirow{3}{*}{\textbf{Method}} & \multicolumn{5}{c}{\textbf{Basic}} \\ \cline{2-6} 
 & Attribute & Scene & \multicolumn{3}{c}{Relation} \\ \cline{4-6} 
 &  &  & Spatial & Action & Part \\ \hline
\textit{Denoised Latent (iteration 21-30)} &  &  &  &  &  \\
~\model & \textbf{0.831} & \textbf{0.854} & \textbf{0.817} & \textbf{0.821} & \textbf{0.820} \\
CLIP L14 & 0.819 & 0.853 & 0.805 & 0.808 & 0.799 \\
LatentCLIP-4-Plus & 0.815 & 0.843 & 0.797 & 0.800 & 0.792 \\
LatentCLIP-8 & 0.811 & 0.851 & 0.794 & 0.798 & 0.785 \\ \hline
\textit{Final Image} &  &  &  &  &  \\
CLIP L14 & 0.846 & 0.862 & 0.837 & 0.838 & 0.842 \\
LatentCLIP-4-Plus & 0.859 & 0.871 & 0.847 & 0.851 & 0.844 \\
LatentCLIP-8 & 0.853 & 0.868 & 0.841 & 0.850 & 0.855 \\ \hline
Mean Value & 0.814 & 0.841 & 0.798 & 0.803 & 0.796 \\ \hline
\end{tabular}
\end{table}

\subsection{Basic Prompts}

Table~\ref{tab:basic_genai_bench} shows that~\model consistently outperforms baselines as well as the mean value of all skills.

Looking more into depth in the Attribute skill, using CLIP to select an image during the generation achieves 0.819 while our approach improves by 1.2\%, showing that our model is capable of selecting images that better depict attributes of entities such as colors and materials while the other finetuned approaches achieve lower scores than CLIP. On the other hand when we go to generations that obligate to focus more on the overall scene such as backgrounds the results are pretty similar between our approach and CLIP being just 0.8\% off of the CLIP choice in the final image. 

Spatial relationship analysis shows that CLIP and LatentCLIP struggle significantly during mid-generation, scoring at least 3.2\% lower to the score on the final image. Our approach, however improves alignment with the second-best by 1.2\%, showcasing its ability to capture physical relations between multiple entities, such as "a cat on the right and a dog on the left". This indicates that our model not only detects entities in noisy conditions, but also captures the differences between them, selecting an image that can achieve closer results to the final scores. Action and Part splits take a similar approach where our model can improve performance of the baselines in the middle of the generation by up to 2.1\%. Notably, despite being trained on latents, LatentCLIP consistently achieves the best results when selecting the final image but performs the worst when applied during the middle of the generation process.

\begin{table}[h]
\centering
\caption{Results of applying the test-time scaling with Best-of-N during the generation process on the advanced prompt of the GenAI-Bench.}
\label{tab:advanced_genai_bench}
\begin{tabular}{lccccc}
\hline
\multirow{3}{*}{\textbf{Method}} & \multicolumn{5}{c}{\textbf{Advanced}} \\ \cline{2-6} 
 & Count & Differ & Compare & \multicolumn{2}{c}{Logical} \\ \cline{5-6} 
 &  &  &  & Negate & Universal \\ \hline
\textit{Denoised Latent (iteration 21-30)} &  &  &  &  &  \\
~\model & \textbf{0.738} & \textbf{0.709} & \textbf{0.721} & 0.517 & \textbf{0.676} \\
CLIP L14 & 0.716 & 0.688 & 0.683 & \textbf{0.522} & 0.674 \\
LatentCLIP-4-Plus & 0.728 & 0.699 & 0.703 & 0.517 & 0.671 \\
LatentCLIP-8 & 0.721 & 0.682 & 0.693 & 0.515 & 0.656 \\ \hline
\textit{Final Image} &  &  &  &  &  \\ 
CLIP L14 & 0.755 & 0.727 & 0.742 & 0.536 & 0.687 \\
LatentCLIP-4-Plus & 0.770 & 0.733 & 0.731 & 0.515 & 0.682 \\
LatentCLIP-8 & 0.765 & 0.733 & 0.743 & 0.531 & 0.683 \\ \hline
Mean Value & 0.717 & 0.694 & 0.698 & 0.523 & 0.657 \\ \hline
\end{tabular}
\end{table}

\subsection{Advanced Prompts}
The results presented in Table~\ref{tab:advanced_genai_bench} demonstrate the effectiveness of our method with "advanced" prompts, highlighting its robust ability to select images that closely match more complex textual descriptions. The lower score for these images accurately reflects the inherent challenge that generative models and twin-tower encoders face when interpreting complex prompts compared to simpler, basic prompts. Interestingly, as prompt difficulty increases, using a randomly selected image can sometimes align better than employing the baselines midway through generation.

Looking at the performance per skill,~\model shows an improvement in selection accuracy for prompts that specify quantity or size. This suggests that our approach helps mitigate the known limitations of CLIP regarding counting, particularly under noisy conditions. For captions requiring differentiation (e.g., "one cat is sleeping on the table and the other is playing under the table"), we observed a further 1 to 2.7\% improvement. This demonstrates that our model can enhance the overall image alignment even with complex spatial or relational descriptions, all while reducing the necessary diffusion steps.

In the comparison skill, our model achieves a gap of 1.8\% to 3.8\% improvement over LatentCLIP and CLIP selection during the generation process, indicating a clear advantage. Consistent with all baselines, our approach struggles most with negation prompts, such as those where entities are missing or actions are unexecuted (e.g., "a shelf without books"). This challenge is common due to the limited negation examples in training datasets, as noted by~\cite{know_no_better}, with no reported method exceeding a score of 54\% even in the final image. Similarly, all approaches struggle with universal attribute sharing, with no method exceeding 70\%.

These results, despite the increased prompt complexity, validate our initial findings: our method remains superior in assessing alignment and selecting the optimal image within the latent space, successfully overcoming challenges from noisy intermediate representations. However, it is important to note that our results also show that known limitations of CLIP, particularly its difficulty with negation, transfer directly to noisy latent space results. In contrast to the results for basic prompts, where CLIP consistently outperformed LatentCLIP mid-generation, the performance between the two baselines is more evenly matched when handling advanced prompts.

\begin{table}[t]
\centering
\label{tab:alignment_vqa}
\caption{Alignment assessed across both GenAI-Bench splits. The table presents the alignment scores for each model, and for the denoised latent output, the alignment is also shown as a percentage of the final image score, offering a thorough perspective on alignment performance.}
\begin{tabular}{@{}lcc@{}}
\toprule
\multirow{2}{*}{Method} & \multicolumn{2}{c}{\textbf{Alignment}} \\ \cmidrule(l){2-3} 
 & Basic & Advanced \\ \midrule
\multicolumn{3}{l}{\textit{Denoised Latent (iteration 21-30)}} \\
\model & \textbf{0.262 (88.5\%)} & \textbf{0.250 (90.6\%)} \\
CLIP L14 & 0.183 (19.8\%) & 0.186 (67.4\%) \\ \midrule
\multicolumn{3}{l}{\textit{Final Image}} \\
CLIP L14 & 0.296 & 0.276 \\ \bottomrule
\end{tabular}
\end{table}

\subsection{Human Evaluation}

To supplement our automated evaluation metrics, we conducted a blind human preference study utilizing 100 randomly sampled image pairs from the GenAI-Bench Basic split. Annotators were tasked with evaluating the images picked by \model versus standard CLIP at latent step 20. Across valid trials (excluding strict ties), human evaluators demonstrated a clear preference for \model, selecting it in 61.3\% of comparisons, backed by strong inter-rater agreement ($\kappa = 0.724$).

\section{Alignment}

In this experiment, we examine the variations in alignment scores produced by~\model and CLIP during the mid-generation phase and compare these results to the score of CLIP on the final output, across both splits of the Noisy-GenAI-Bench.

Specifically, with basic prompts, \model reaches 88.5\% of the final score during the mid-generation stage, significantly outperforming the CLIP baseline, which only achieves 19.8\%. In contrast, when using advanced prompts, \model achieves 90.6\% of the final score, while CLIP baseline achieves 67.4\%. Our approach shows a major gain over the baseline, indicating that fine-tuning on noisy latent data can substantially improve similarity measurement in the middle of the generation, achieving over 90\% of the final score.

\section{Ablation Studies}
\subsection{Impact of Trained Encoders}

\begin{table}[t]
\centering
\caption{Comparative analysis of performance improvements when training only the image encoder compared to joint training with the text encoder.}
\label{tab:encoder-results}
\begin{tabular}{lcc}
\toprule
\multirow{2}{*}{Model} & \multicolumn{2}{c}{\textbf{GenAI}} \\
\cmidrule(lr){2-3}
 &  Basic & Advanced \\
\midrule
\multicolumn{3}{@{}l}{\textit{Denoised Latent (iteration 21-30)}} \\
Frozen & 0.821 & 0.633 \\
+ Image Encoder & \textbf{0.836} & \textbf{0.646} \\
+ Text Encoder & 0.835 & 0.645\\
\midrule
\multicolumn{3}{@{}l}{\textit{Image Dimension}} \\
Frozen & 0.856 & 0.664 \\
+ Image Encoder & \textbf{0.858} & \textbf{0.664}\\
+ Text Encoder & 0.856 &  0.657\\
\bottomrule
\end{tabular}
\end{table}

To evaluate how the trained modalities influence the perfomance of~\model, we report experiments that evaluate three approaches: a frozen baseline, training the image encoder, and a fully trained text and image encoder on the Noisy-GenAI-Bench. As illustrated in Table~\ref{tab:encoder-results}, both training approaches show enhancements in both subsets. In particular, training only the image encoder consistently delivers the best results. Within the latent dimension, the differences between the image-trained model and the fully trained model are relatively small, while the difference compared to the frozen baseline is more significant (up to 1.3\%). However, in the image dimension, the text encoder demonstrated a worse performance than the frozen baseline, particularly within the ‘Advanced’ subset, while the trained image encoder maintained comparable or superior results.

The study shows that training the image encoder is critical for performance, offering up to 1.5\% gains over the baseline, confirming our choice to prioritize image encoder optimization to increase the capabilities of the twin-tower encoder under noise.

\subsection{Analysis on Scaling Model Size}

To evaluate the correlation between model size and performance, we compared~\model using two different encoder sizes, ViT-B/16 (150M parameters) and ViT-L/14. These results are presented in Table~\ref{tab:model_sizes}, focusing on performance in the Basic and Advanced subsets of Noisy-GenAI-Bench.
The results consistently show a positive correlation between model size and performance when fine-tuned. The larger ViT-L/14 encoder generally surpasses the smaller ViT-B/16 across both evaluation settings. In the latent dimension, ViT-L/14 outperformed ViT-B/16 by 0.7\% on the Basic set and 0.5\% on the Advanced set. The advantage was more pronounced in the Image Dimension setting, where ViT-L/14 scored 0.858 and 0.664, surpassing ViT-B/16 scores (0.854 and 0.652) by 0.5\% and 1.8\%, respectively. This ablation study validates that scaling the encoder size is an effective strategy to improve performance even in this noisy environment, suggesting that models with higher parameter counts possess a greater capacity to learn robust representations. With this we cannot discard the fact that the relatively small difference in final scores indicates that even the smaller ViT-B/16 encoder achieves very competitive results.

\begin{table}[h]
\centering
\setlength{\tabcolsep}{4pt}
\caption{Analysis of performance improvements by varying the encoder size (Evaluated at Latent Dimension Steps 21-30)}
\label{tab:model_sizes}
\begin{tabular}{@{}llcc@{}}
\toprule
\textbf{Encoder} & \textbf{Encoder Size} & \multicolumn{2}{c}{\textbf{GenAI Score}} \\ \cmidrule(l){2-3} \cmidrule(l){3-4} 
Size & (Parameters) & Basic & Advanced \\ \midrule
\multicolumn{4}{l}{Denoised Latent (iteration 21-30)} \\
ViT-B/16 & 150M & 0.829 & 0.643 \\
ViT-L/14 & 428M & \textbf{0.836} & \textbf{0.646} \\ \midrule
\multicolumn{4}{l}{Image Dimension} \\
ViT-B/16 & 150M & 0.854 & 0.652 \\
ViT-L/14 & 428M & \textbf{0.858} & \textbf{0.664} \\ \bottomrule
\end{tabular}
\end{table}

\section{Further Analysis of Qualitative Results}

\begin{figure*}[ht]
    \centering
    \begin{subfigure}[b]{\linewidth}
        \centering
        \includegraphics[width=\linewidth]{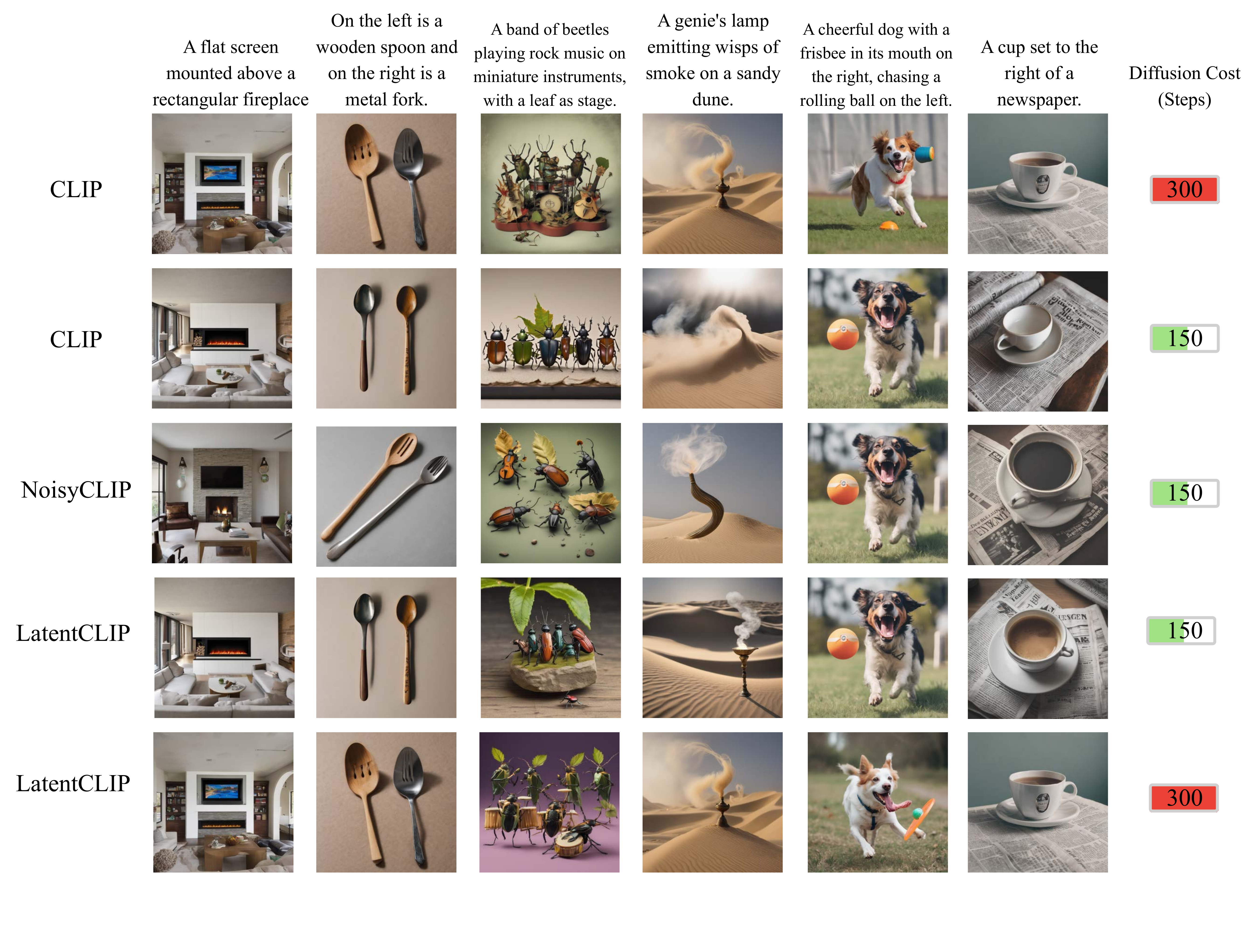}
        \label{fig:results_1_sub1}
    \end{subfigure}
    \vspace{-30pt}
    \begin{subfigure}[b]{\linewidth}
        \centering
        \includegraphics[width=\linewidth]{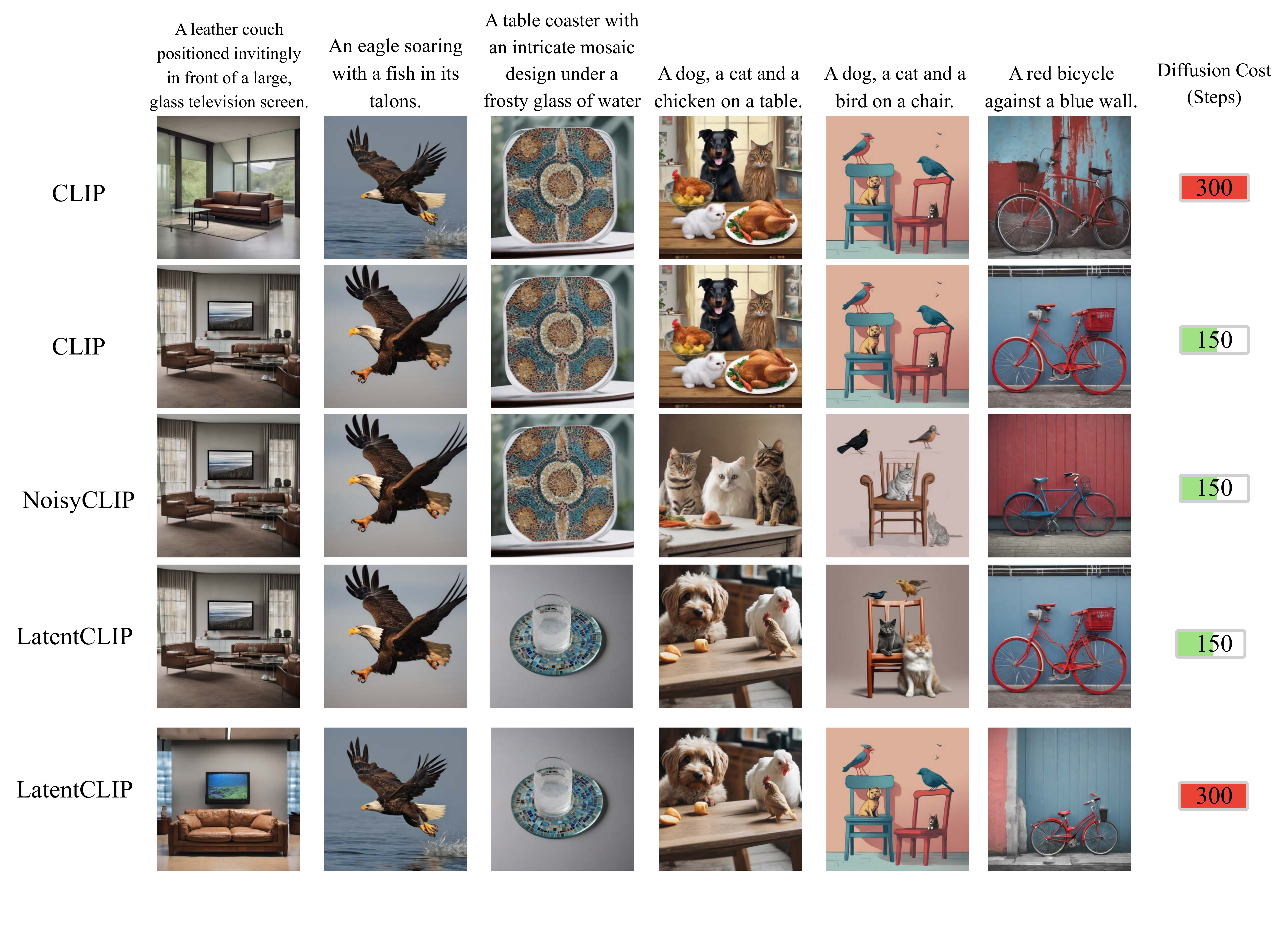}
        \label{fig:results_1_sub2}
    \end{subfigure}
    \caption{Images are generated using a best-of-6 strategy. Our proposed method is compared against CLIP and LatentCLIP at two computation budgets: a fixed lower cost and the full cost (generating all 6 images).}
    \label{fig:results_appendix}
\end{figure*}

\subsection{Analysis of Edge Scenarios on Best-of-N}

This section extends the qualitative results by presenting an analysis of borderline cases and edge scenarios in the task of image selection during and after the generation process.

In Figure~\ref{fig:results_appendix} (top), the first example demonstrates~\model's ability to select an image with alignment similar to post-generation baselines, whereas CLIP and LatentCLIP applied during generation lacks overall compositional quality. The second example, despite the poor generated images, shows \model selecting the closest match to the prompt during the generation, while CLIP and LatentCLIP incorrectly reverse the order of the utensils. The third prompt highlights that with more complex prompts none of the models were capable of selecting an aligned image with half of the computation.

However, the fourth example shows a failure case for~\model and CLIP where they select what appears to be a hallucination. This likely occurred because the generation model failed to produce the requested object and instead focused on an unintended entity (the genie), which~\model and CLIP mistakenly prioritized.  In the fifth case, where no single image depicts both the ball and the frisbee, different methods focus on one or the other; only LatentCLIP selects an image that partially aligns with the prompt, though this requires running the generation to completion. Similarly, the final prompt highlights a scenario where all methods fail to produce a perfect match, reflecting the inherent limitations of the underlying generative process.

In the next set of examples (Figure~\ref{fig:results_appendix} (bottom), the second example again depicts a scenario where no image was correctly generated, resulting in all methods failing to select an accurate representation of the prompt. The third prompt highlights a failure case, where CLIP and~\model over-focused on the initial part of the prompt while omitting the important "frosty glass of water".

The fourth and fifth prompts illustrate cases where all models exhibit hallucinations regarding the count of entities. Specifically, in the fifth prompt, CLIP selects an image featuring the three required animals (with two birds instead of one) but incorrectly includes two chairs instead of the one specified. In contrast, our approach selects an image that correctly depicts a single chair as prompted, but entirely misses the dog and contains incorrect counts of other entities. LatenCLIP shares the same problems by choosing image with a single chair and without a dog during the middle of the generation and chaging in the end choosing the same image as CLIP.  This strong divergence suggests that with complex, multi-entity prompts, selection choices can highly vary based on the prioritization of different attributes. Finally, the sixth prompt represents a failure case for our model, highlighting its susceptibility to misunderstanding or swapping the attributes of different elements.

\subsection{Examples of Middle Diffusion Scoring}
\begin{figure}[h]
    \centering
    \includegraphics[width=\linewidth]{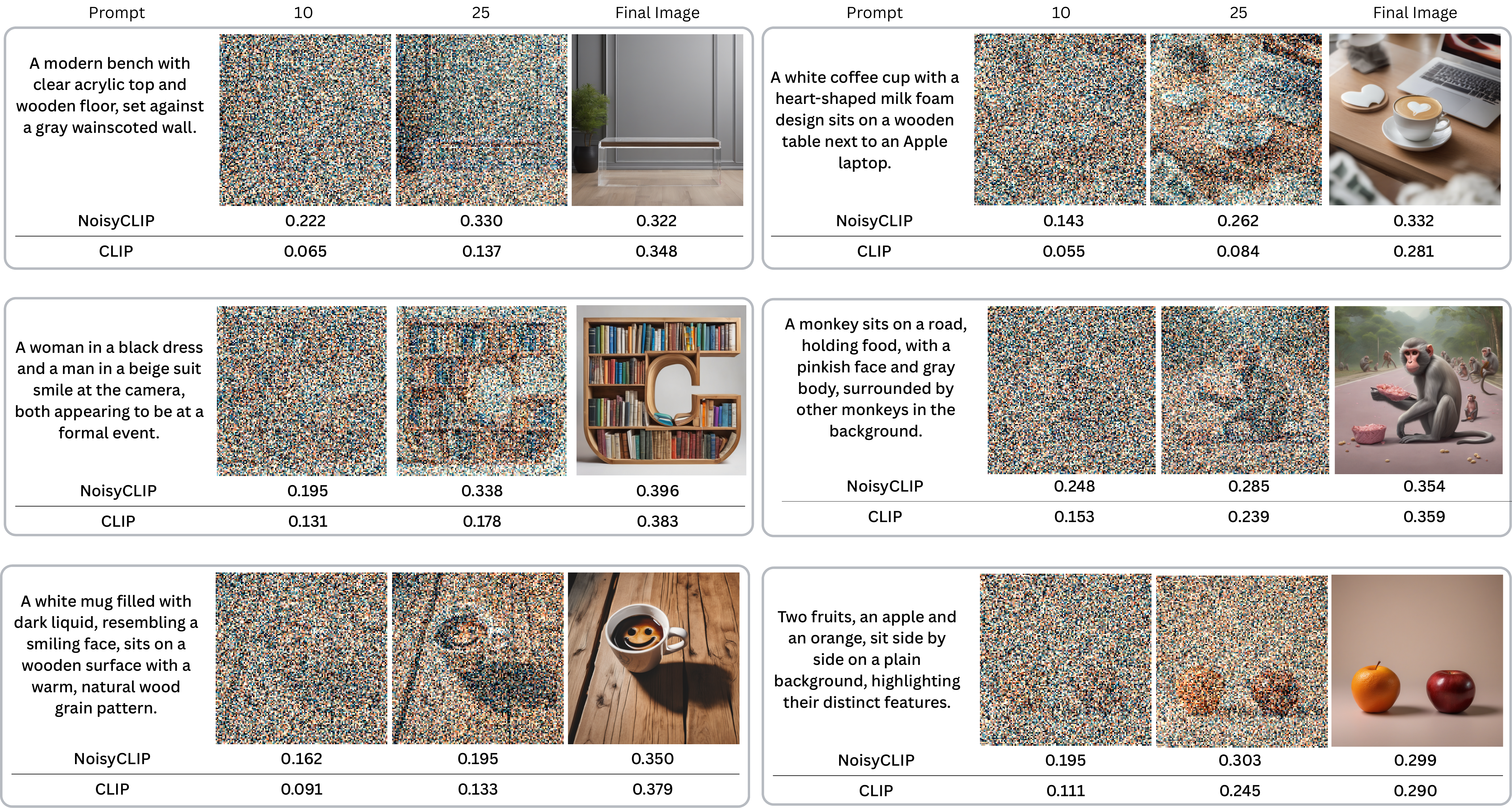}
    \caption{Image generation process using \model and CLIP for early alignment assessment during generation.}
    \label{fig:results_with_scores}
\end{figure}

To conclude we present results illustrating the behaviour of the scoring functions during the reverse diffusion process (Figure~\ref{fig:results_with_scores}), where we compare our model with the backbone. Each example shows a prompt and three generation stages, an early noisy latent (step 10), a middle-point latent (step 25), and the final image. We measured prompt-image similarity using both~\model and CLIP in each depicted stage. Overall, our model consistently achieves scores in the middle of the generation that closely correlate with the final scores. This demonstrates how our approach enables twin-tower encoders to effectively measure similarity under noisy conditions, allowing for an accurate assessment of the promp-to-image alignment mid-generation.

\end{document}